\documentclass[]{1181}
\usepackage{lipsum} 
\usepackage[hidelinks]{hyperref}
\pdfoutput=1

\usepackage{epstopdf}
\usepackage[caption=false]{subfig}

\usepackage[longnamesfirst,sort]{natbib}
\bibpunct[, ]{(}{)}{;}{a}{,}{,}
\renewcommand\bibfont{\fontsize{10}{12}\selectfont}

\theoremstyle{plain}

\theoremstyle{definition}

\theoremstyle{remark}

\begin{document}

\title{Development of a Risk-Free COVID-19 Screening Algorithm from Routine Blood Tests Using Ensemble Machine Learning}

\author{
\name{Md. Mohsin Sarker Raihan\textsuperscript{a} \thanks{CONTACT Md. Mohsin Sarker Raihan. Email: msr.raihan@gmail.com}, 
      Md. Mohi Uddin Khan\textsuperscript{b}, 
      Laboni Akter\textsuperscript{c} and 
      Abdullah Bin Shams\textsuperscript{d}}
\vspace{7pt}
\affil{Department of Biomedical Engineering, Khulna University of Engineering \& Technology, Khulna, Bangladesh\textsuperscript{a,c};\\ Department of Electrical \& Electronic Engineering, Islamic University of Technology, Boardbazar, Gazipur, Bangladesh\textsuperscript{b};\\ Department of Electrical \& Computer Engineering, University of Toronto, Toronto, Ontario, Canada.\textsuperscript{d}}
\vspace{7pt}
\email{\href{mailto:msr.raihan@gmail.com}{msr.raihan@gmail.com}\textsuperscript{a}; 
       \href{mailto:mohiuddinkhan5252@gmail.com}{mohiuddinkhan5252@gmail.com}\textsuperscript{b}; \href{mailto:laboni.kuet.bme@gmail.com}{laboni.kuet.bme@gmail.com}\textsuperscript{c}; 
       \href{mailto:abdullahbinshams@gmail.com}{abdullahbinshams@gmail.com}\textsuperscript{d}}
}

\maketitle

\begin{abstract}
The Reverse Transcription Polymerase Chain Reaction (RTPCR) test is the silver bullet diagnostic test to discern COVID infection. Rapid antigen detection is a screening test to identify COVID positive patients in little as 15 minutes, but has a lower sensitivity than the PCR tests. Besides having multiple standardized test kits, many people are getting infected and either recovering or dying even before the test due to the shortage and cost of kits, lack of indispensable specialists and labs, time-consuming result compared to bulk population especially in developing and underdeveloped countries. Intrigued by the parametric deviations in immunological and hematological profile of a COVID patient, this research work leveraged the concept of COVID-19 detection by proposing a risk-free and highly accurate Stacked Ensemble Machine Learning model to identify a COVID patient from communally available-widespread-cheap routine blood tests which gives a promising accuracy, precision, recall and F1-score of 100\%. Analysis from R-curve also shows the preciseness of the risk-free model to be implemented. The proposed method has the potential for large scale ubiquitous low-cost screening application. This can add an extra layer of protection in keeping the number of infected cases to a minimum and control the pandemic by identifying asymptomatic or pre-symptomatic people early.
\end{abstract}


\section{Introduction}
The pandemic by SARS-CoV-2 infection has claimed over 4 million lives around the world to this date. Since the outbreak, countries have expeditiously ramped up their capacity in full for testing and contract tracing, emanating a 195.545 million confirmed cases from 2.35 billion tests conducted worldwide \citep{1}.  The sub-atomic experiment accomplished using the converse polymerase chain response (PCR) method is the tool of choice, or the best quality standard, for detecting SARS-CoV-2 contamination. However, the test is time-consuming, necessitating the use of specialized hardware and substances, the collaboration of particular and qualified recruits designed for the sample assortment, and relying on the sufficient hereditary security of the RNA groupings chosen for tempering the preliminary \citep{2}. Attempt to develop tests based on IgM/IgG antibodies also suffer from low sensitivity, specificity and costly reagents.

To improve symptomatic capacities, the information science local area has suggested a few AI machine learning (ML) models. The majority of these models depend on processed tomography outputs or chest X-ray beams. Despite the detailed promising outcomes, a few concerns have been higher in regards to these and different works, particularly as to arrangements dependent on chest X-ray beams, which have been related with high number of negative outcomes. Then again, arrangements dependent on CT imaging are influenced by the attributes of this methodology: CTs are exorbitant, tedious, and need particular devices; accordingly, methodologies dependent on this imaging procedure cannot be applied sensibly for screening tests. Detection based on chest CT images and X-rays cannot be performed abundantly due to even higher cost \& radiation exposure.

Considering the detrimental effects of nCoV breeding inside host lung cells utilizing the Angiotensin-Converting Enzyme 2 (ACE2) found in ample amount on type-II alveolar cells and considering inflammation in lungs \& respiratory tracts, recent studies have reported significant change in immunological \& hematological parameters in the host blood stream. The predominant purpose of this study is to develop a novel machine learning technique, featuring diagnosis of COVID-19 subjects that have undergone routine pathological blood tests.

This study proposes a Double Layer Stacked Ensemble Machine Learning model that can classify the individuals whether they are infected with nCoV or not with high accuracy. Despite focusing on higher accuracy, this study also gave in equal importance to achieve high precision \& recall so that no individual is misclassified. The unconventionality of this study is that the research has improved the overall performance metrics of the proposed model by analyzing consequences of misclassification based on different Basic Reproduction Numbers $(R_{0})$ related to COVID-19 pandemic. The proposed model takes an attempt to improve COVID-19 patient classification ML algorithm so that a cheap and widely available COVID diagnostic method can be developed utilizing blood samples.

\section{Related Works}
AlJame et al.~\citep{3} flourished ML model named ERLX for identifying COVID-19 from routine blood tests. There, two levels had been used wherein first level included Random Forest, Logistic Regression, and Extra Trees \& XGBoost classifier in another level with 18 features. KNNImputer algorithm had been used to manage null values in the dataset, iForest used to eliminate outlier data, and SMOTE technique used to overcome an imbalanced data; feature importance is described by using the SHAP system. In that study, sample data was 5644 where 559 was confirmed COVID-19 circumstances. The model obtained the result with average accuracy of 99\%, AUC of 99\%, a sensitivity of 98\%, and a specificity of 99\%.

Barbosa et al.~\citep{6} proposed a framework that helps Covid-19 determination dependent on blood testing. In this study, the laboratory parameters achieved 24 features from biochemical and the hemogram tests characterized to help clinical analysis. They used several types of machine learning techniques namely Random Forest, SVM, Bayesian Network, Naïve Bayes and achieved the accuracy 95.2\%, sensitivity  96.8\%, kappa index 90.3\%, specificity 93.6\%, and precision 93.8\%.

Kukar et al.~\citep{7} represented a method to predict COVID-19 infection via ML dependent on clinically accessible blood test results. Using the Random Forest algorithm with 24 features, the COVID-19 screening achieved an in-depth clinical evaluation with accuracy of 91.67\%.

\section{Methodology}
Blood test in healthcare is performed in order to ascertain the Biochemical \& Physiological state of a person. The whole dataset containing various attributes comprising hemocyte count \& hematochemical profile obtained from two types of blood-test, viz. Complete Blood Count (CBC) \& Basic Metabolic Panel, is analyzed according to the block diagram portrayed in \figureautorefname{-3.1}. 

\begin{figure}[h]
\renewcommand{\thefigure}{3.1}
\centering
\resizebox*{10cm}{!}{\includegraphics{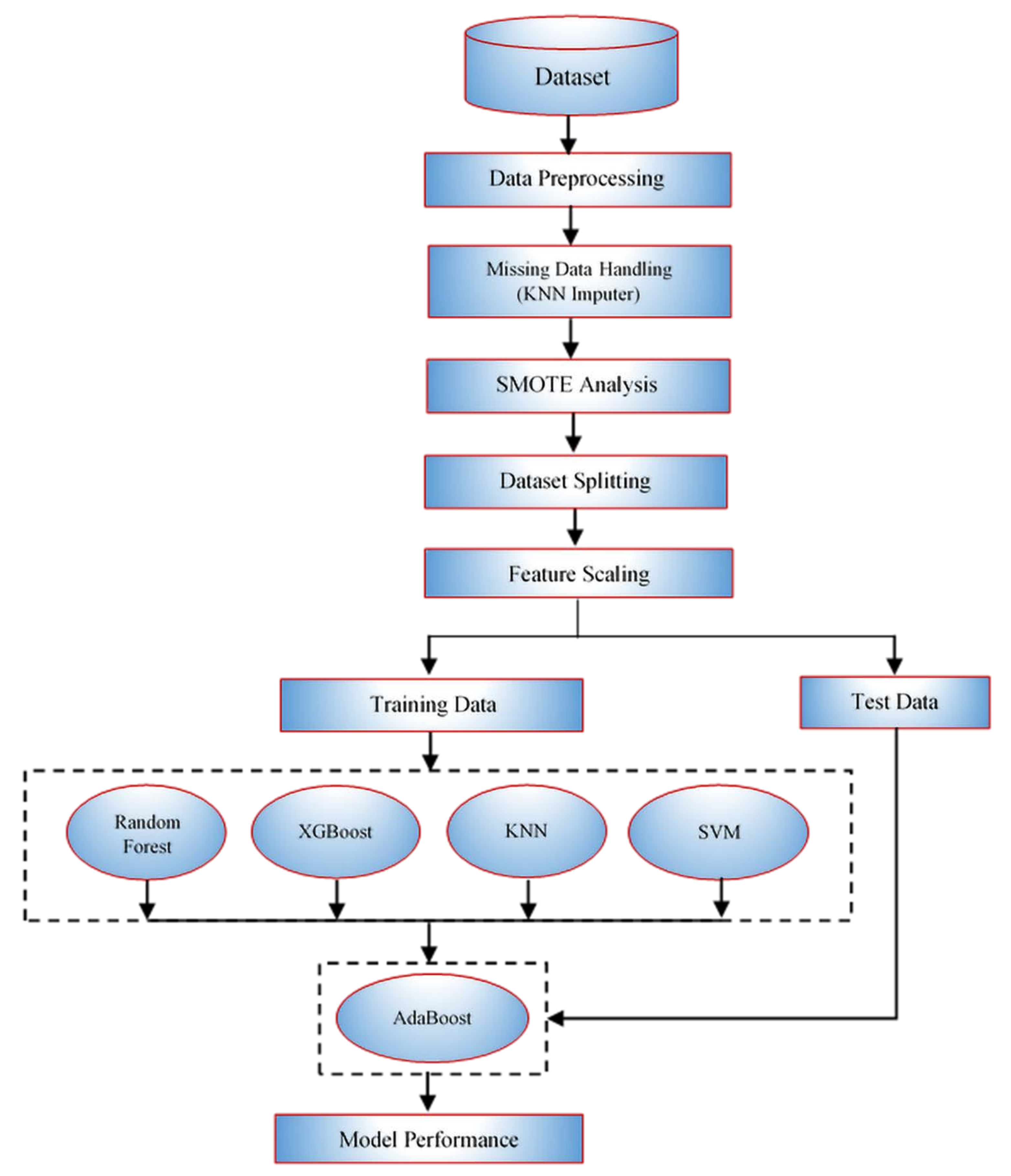}}\hspace{5pt}
\caption{Block diagram describing the proposed method for the classification of COVID-19 based on Blood Sample data.}
\label{fig_3.1}
\end{figure}

\subsection{Dataset Collection}
The dataset used in this research was collected from Kaggle \citep{k} which includes 5644 samples with 111 attributes. Among these, 603 samples \& only blood related 25 attributes were selected for this study. The attributes are: Hematocrit, Hemoglobin, Platelets, MPV, RBC, Lymphocytes, MCHC, Leukocytes, Basophils, MCH, Eosinophils, MCV, Monocytes, RDW, Neutrophils, Potassium, Creatinine, Sodium, Aspartate, Transaminase, INR, Albumin, Alanine, Transaminase, Proteina C reativa.

\subsection{Data Pre-processing}

\subsubsection{Missing Data Handling}
Several techniques are available to counter missing values in a dataset. In this research work, the KNN imputer has been applied with k=5 number of neighbors to eliminate missing data contained in the dataset~\citep{3}.

\subsubsection{SMOTE Analysis}
An imbalanced dataset makes the classifier model highly biased towards the high frequency target class. To balance data, a well-known oversampling technique named `Synthetic Minority Over-sampling Technique (SMOTE)' has been used in this study which does data augmentation by grabbing the concept of k-nearest neighbors. Augmented data is intelligently added to be used in the classification problem to improve the data distribution. SMOTE does not augment data freely but similar to the existing one~\citep{11}.

\begin{figure}[h]
\renewcommand{\thefigure}{3.2.2.1}
\centering
\resizebox*{7cm}{!}{\includegraphics{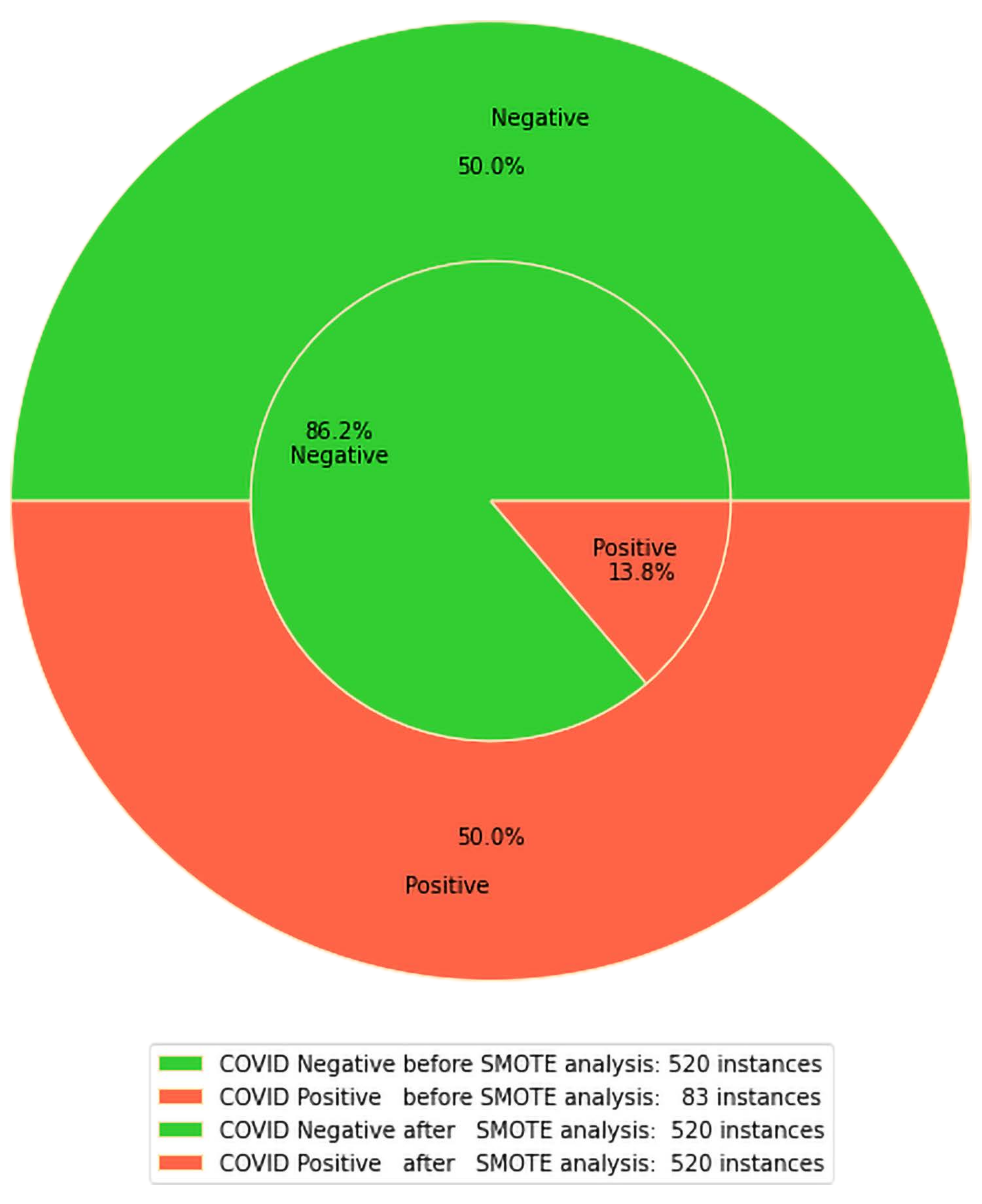}}\hspace{5pt}
\caption{Data distribution before (inner-ring) \& after (outer-ring) SMOTE analysis.}
\label{fig_3.2.2.1}
\end{figure}

\subsubsection{Data Splitting}
The dataset was splitted into 70:30 ratio as training \& test set. The stacked ensemble model was trained using 70\% training data \&  overall system evaluation was performed using 30\% test data. 

\subsubsection{Feature Scaling}
In this study, standardization has been used as the feature scaling method. The esteems are located near the average by a unit standard deviation in such a scaling process. This senses that the feature's average turns into zero, then the distribution takes a unit standard deviation after using this technique~\citep{12}.

\subsection{Stacked Ensemble Machine Learning }
This technique has widely been used since invention in which the $1^{st}$ level learners are trained \& utilized to make primary prediction. The $1^{st}$ level predictions are combined as the training dataset for a stacked $2^{nd}$ level learner called meta-learner \& the test dataset is then fed to $2^{nd}$ level learner in order to make final prediction~\citep{12a}.

\subsection{Machine Learning Algorithms }
This study maneuvers double layer stacked ensemble machine learning in order to precisely classify the subjects under Severe Acute Respiratory Syndrome (SARS) test caused by SARS-CoV-2 strain of coronavirus as COVID positive or negative based on the hematological profile containing 25 attributes of 1040 subjects. Four algorithms (K-Nearest Neighbors, Support Vector Machine, Random Forest, XGBoost) were chosen for the $1^{st}$ layer \& AdaBoost was chosen as meta\-learner for the $2^{nd}$ layer of the proposed model based on high accuracy, precision, recall \& F1-score.

\subsubsection{K-Nearest Neighbors (KNN) }
This algorithm measures the distance of the test datapoint from the training datapoints \& classifies the test datapoint according to the class of the k-closest training data-points in the neighborhood \citep{13}.

The hyperparameters this study used for KNN: 
{\bf n\_neighbors $= 3$, metric $=minkowski $, p  $=2$}.

\subsubsection{Support Vector Machine (SVM) }
This algorithm draws the best decision boundary called ‘hyperplane’ at maximum distances from the target class in n-dimensional space so that new data points can certainly be categorized into the right classification later on \citep{13}.

The hyperparameters this study used for SVM: 
{\bf kernel $=rbf $, random\_state $= 0$}.

\subsubsection{Random Forest (RF)}
To overcome pitfalls (data overfitting) of Decision Tree (DT), RF produces a series of Decision Trees that have been trained using the `bagging' technique wherein the general result is built from a multitude of DT learning models \citep{13}. 

The hyperparameters this study used for RF: 
{\bf n\_estimators $= 49$, criterion $= entropy $, random\_state $= 0$}.

\subsubsection{XGBoost (XGB) }
It’s an optimized distributed gradient boosting library that utilizes ensemble learning techniques. Parallel \& distributed computing from an ensemble of various learning algorithms makes it faster \& efficient \citep{14}.

The hyperparameters this study used for XGB: 
{\bf n\_estimators $= 83$, max\_depth $=12$, subsample $=0.7$}.

\subsubsection{AdaBoost (AdB) }
Boosting algorithms deal with both bias-variance trade-off while bagging algorithms control variance only. The tweak behind improved accuracy \& performance of AdB is the combination of weighted sum of few weak-learning classifiers that were misclassifying some instances when implied individually \citep{14}.

The hyperparameters this study used for AdB: 
{\bf n\_estimators $=67$, learning\_rate $=1$}.

\subsection{Compute Statistical Metrics}
The following statistical metrics were being used to measure the performance of this proposed work: accuracy, precision, recall, and f1 score. Individual output is calculated using confusion matrix values such as true negative (TN), true positive (TP), false negative (FN), and false positive (FP) \citep{15}.

\section{Result and Discussion}
The performance metrics of the detailed study are tabulated in \tableautorefname{-4.1} \& graphically rendered in \figureautorefname{-4.1}. 

\begin{table}[h]
\renewcommand{\thetable}{4.1}
\tbl{Performance Metrics of the Proposed Stacked Ensemble Model.}
{\begin{tabular}{cccccccc} \toprule
 & & \multicolumn{4}{c}{\bf{Layer-1}} & &  \bf{Layer-2}  \\ \cmidrule{3-6} \cmidrule{8-8}
\bf{Performance Metrics} & &  KNN  &  SVM  &  XGB  &  RF  & &  AdB \\ \midrule
Accuracy   & &  $84.62\%$  &  $92.95\%$  &  $95.19\%$  &  $97.12\%$  & &  $100\%$  \\
Precision  & &  $98.73\%$  &  $96.84\%$  &  $97.47\%$  &  $99.37\%$  & &  $100\%$  \\
Recall  & &  $77.23\%$  &  $90.00\%$  &  $93.33\%$  &  $95.15\%$  & &  $100\%$  \\
F1 Score  & &  $86.67\%$  &  $93.29\%$  &  $95.36\%$  &  $97.21\%$  & &  $100\%$  \\ \bottomrule
\end{tabular}}
\label{table_4.1}
\end{table}

\begin{figure}[h]
\renewcommand{\thefigure}{4.1}
\centering
\resizebox*{8.5cm}{!}{\includegraphics{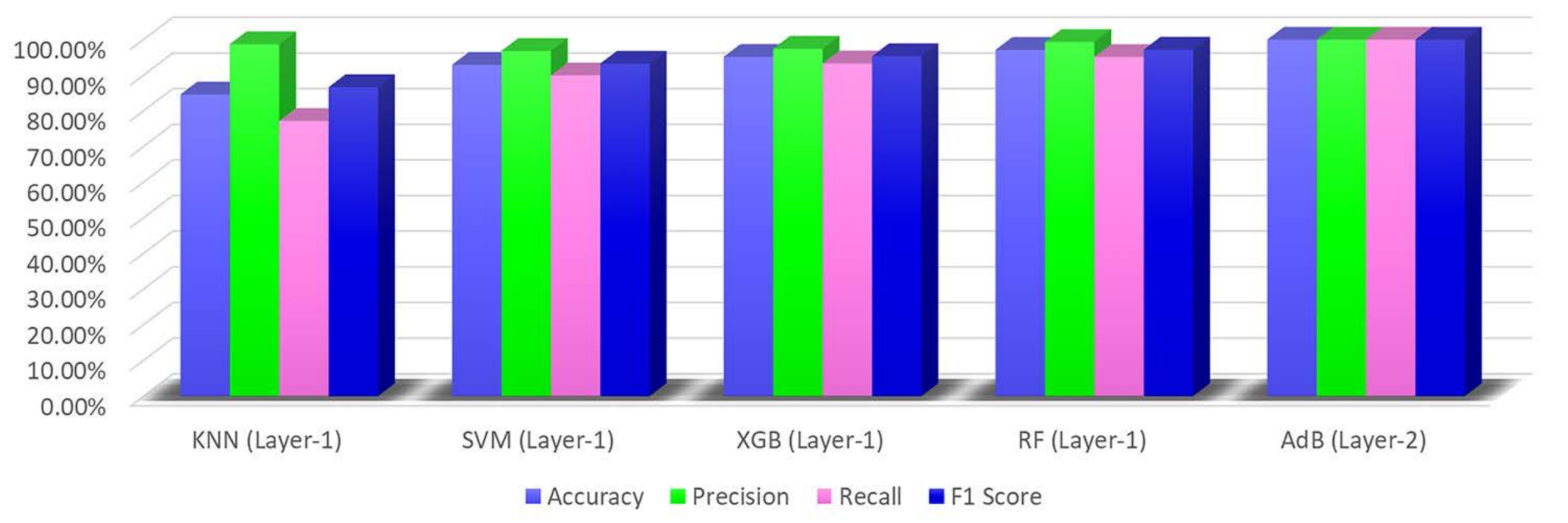}}\hspace{5pt}
\caption{Performance Metrics (Graphical Visualization) of the Proposed Stacked Ensemble Model.}
\label{fig_4.1}
\end{figure}

At $1^{st}$ level classifier, analysis was carried out using different machine learning algorithms (LR, KNN, DT, ANN, SVM, XGB, RF, NB \& AdB) in an effort to find out the best suited algorithms for the $1^{st}$ layer of the proposed model. Based on `accuracy, precision, recall, F1-score' \& giving priority on `precision', four algorithms were chosen for the $1^{st}$ layer of the proposed model which are RF, XGB, SVM \& KNN.

RF, XGB, SVM \& KNN algorithms showed the accuracy of 84.62\%, 92.95\%, 95.19\% \& 97.12\%; the precision of 98.73\%, 96.84\%, 97.47\% \& 99.37\%; the recall of 77.23\%, 90\%, 93.33\% \& 95.15\%, and the F1-score of 86.67\%, 93.29\%, 95.36\% \& 97.21\% respectively. 

It’s worth notifying that despite better performance, none of the algorithms in $1^{st}$ layer are reasonably precise to curb the community transmission of COVID-19.

For illustration of the associated risks, let’s consider the confusion matrix (\figureautorefname{-4.2(a)}) of the best performing algorithm (RF) of the $1^{st}$ layer.

\begin{figure}[h]
\renewcommand{\thefigure}{4.2}
\centering
\resizebox*{8cm}{!}{\includegraphics{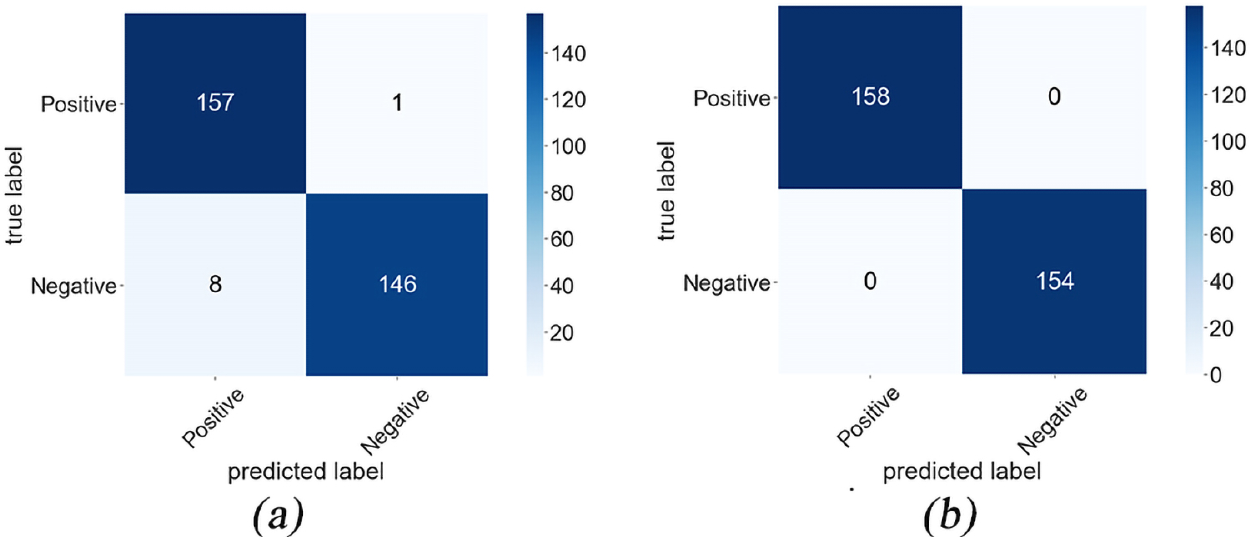}}\hspace{5pt}
\caption{Confusion matrices of the best algorithm   (a) Random Forest in $1^{st}$ layer    (b) AdaBoost in $2^{nd}$ layer.}        
\label{fig_4.2}
\end{figure}

In the $1^{st}$ layer RF, corresponding ‘precision’ is 99.37\% \& ‘recall’ is 95.15\%. Note that precision unfolds the caliber of a model to classify positive values correctly \& recall tells ‘how often the model actually predicts the correct positive values’ \citep{15}.

It’s clear from the confusion matrix of \figureautorefname{-4.2(a)} that the number of people who are actually COVID negative but misclassified as positive (False Positive) is 8. If these FP individuals are now kept in Quarantine/Isolation, they will lose working hours ultimately hampering their monthly income \& the country’s GDP growth. In the worst-case scenario, if they are medicated for COVID-19, it’s left for the reader to think about the consequence of the false medication if the individuals had other chronic diseases.

Again, the dangerous \& risky scenario of the confusion matrix of \figureautorefname{-4.2(a)}, the number of people who are actually COVID positive but misclassified as negative (False Negative) is 1. This ‘False Negative’ classification is of utmost importance because these misclassified people are enough to spread COVID through community transmission if they move freely being assured of negative from COVID test which is wrong.

According to Simple Infectious Disease Model \citep{16}, after the incubation period $\tau$ days, if an infected person, infects exactly $R_{0}$ new susceptible, then the number of new infected individuals only from him/her is given as:
\begin{equation}
n_{E}(t) = n_{E}(0)R_{0}^{{t/\tau }}
\end{equation}
Considering 1 initial infected (misclassified as negative) person \& 7 days incubation period, the equation is graphically visualized in \figureautorefname{-4.3(a)} describing one month projection of new infection cases by only 1 person due to community transmission of COVID for various $R_{0}$ values. It will be seldom possible to curb the pandemic if the newly infected persons spread the disease again like \figureautorefname{-4.3(c)}.

\begin{figure}[h]
\renewcommand{\thefigure}{4.3}
\centering
\resizebox*{13cm}{!}{\includegraphics{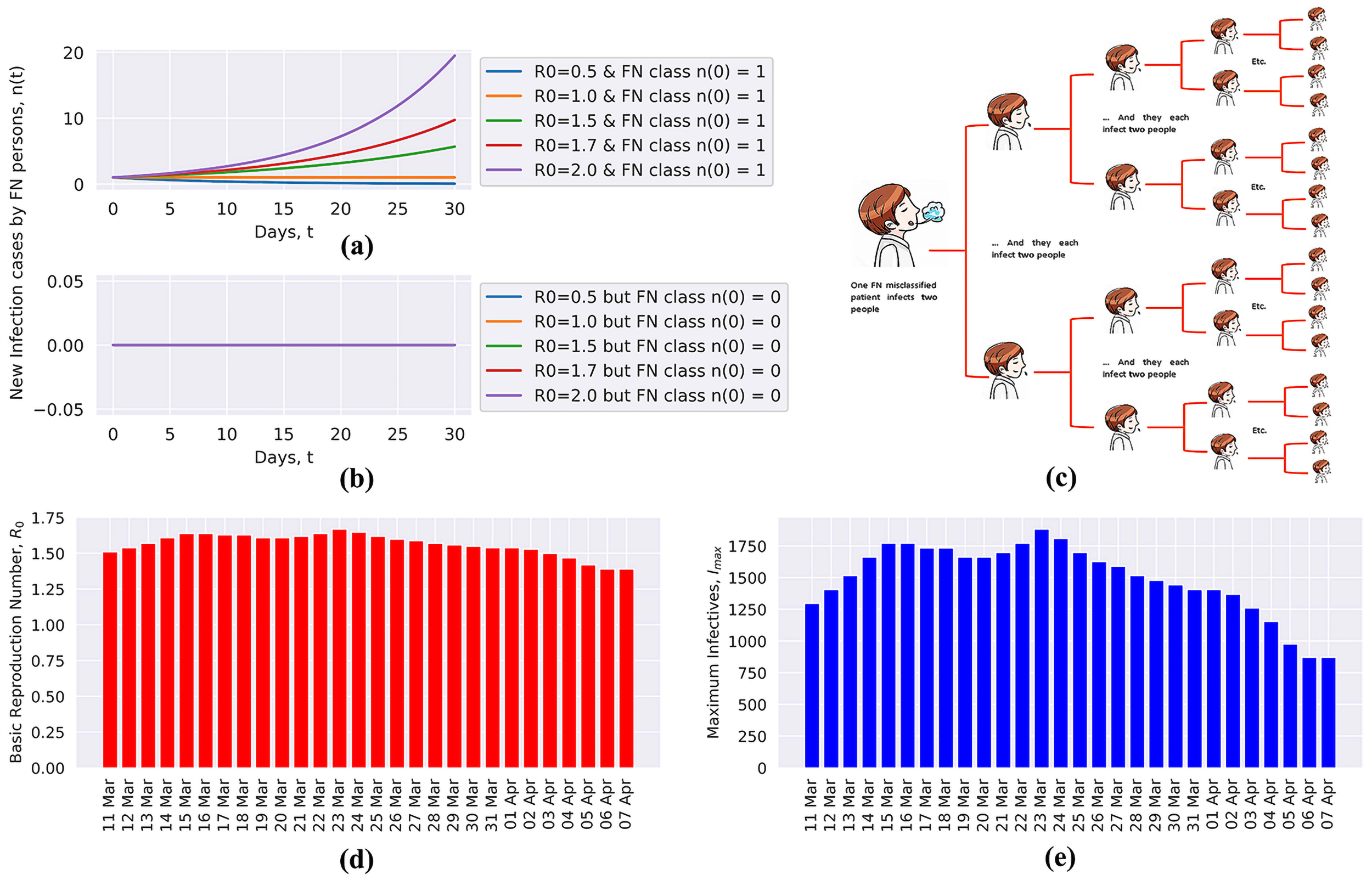}}\hspace{5pt}
\caption{ (a) R-curve: Effect of FN misclassification based on different $R_{0}$ values according to Simple Infectious Disease Model. 
		(b) R-curve remains at zero if the individuals were properly classified.
		(c) Rapid contagious nature of infectious disease if $R_{0}$ = 2.
		(d) COVID-19 $R_{0}$ values of Bangladesh during 11 March 2021 to 7 April 2021.
		(e) Using $R_{0}$ values, rough idea of new infection cases as per simple SIR model.
	}
\label{fig_4.3}
\end{figure}

Now, to quantify the risk regarding infectious disease spread like \figureautorefname{-4.3(c)}, let an area having 20,000 population, where initial infectives (misclassified as COVID negative), $I_{0}$ = 1 \& remaining susceptible, $S_{0}$ = 19,999. According to the simple SIR model for COVID-19 \citep{17}:

Contact ratio,  \( q =\frac{R_{0}}{S_{0}}\)  \&   Maximum number of people that will be infected, \(I_{max}=I_{0}+S_{0}-\frac{1}{q}(1+ln(qS_{0}))\).
Using recent COVID Reproduction Number ($R_{0}$) of Bangladesh \figureautorefname{-4.3(d)} \citep{18} \& equations above, \figureautorefname{-4.3(e)} depicts approximation of maximum number of new infection cases.

This study used a relatively small test set in the $1^{st}$ layer where the best algorithm gave FN $I_{0}$ = 1. In case of larger datasets of bigger area, this algorithm would contribute higher FN $I_{0}$ to the equation of $I_{max}$ \& consequently \figureautorefname{-4.3(e)} would become worse.

If the machine learning algorithms with high accuracy but unknown/less precision \& recall of the reviewed literature (\tableautorefname{-4.2}) keep performing FN misclassification due to less precision \& recall, \figureautorefname{-4.3} clearly says, the \textit{pandemic} will remain a \textit{pandemic} due to community transmission from the FN class.

Therefore, the novelty of the idea \& algorithm of this study is, it puts a shade on the risk factor associated with less precision \& recall using R-curve plot \& $I_{max}$ plot from SIR model using simplified-generalized equations. It is imperative to resolve the FN \& FP classification for which this research work emphasized on improving precision \& recall along with achieving high accuracy by feeding the output of the 4 selected algorithms of $1^{st}$ layer of the proposed model to the $2^{nd}$ (final) layer as input.

In the $2^{nd}$ (final) layer called the meta-learner of the stacked ensemble model, after applying AdaBoost algorithm, the model achieved an improved performance (\tableautorefname{-4.1} \& \figureautorefname{-4.1}) where the AdB algorithm outperformed amongst others achieving all the performance metrics as 100\%.
 
Visually, the confusion matrix \figureautorefname{-4.2(b)} of the best algorithms (AdB) of the $2^{nd}$ layer depicts there is no FN \& FP classification.

Hence, according to the R-curve of \figureautorefname{-4.3(b)}, the new infection cases by FN class will be zero irrespective of different $R_{0}$ values. Consequently, if the true positive individuals are held in isolation \& all social norms are maintained, gradually the $R_{0}$ number will go below 1 \& new infection cases will reduce like \figureautorefname{-4.3(a)} where \(R_{0}<1\).

Also due to zero FP class, the true COVID negative individuals will get relief from: being misclassified, losing working hours \& from taking false medication in worst cases.

\hspace{2pt}
\begin{table}[h]
\renewcommand{\thetable}{4.2}
\tbl{Comparision of the proposed model with other existing models.}
{\begin{tabular}{cclcccc} \toprule
 \bf{References}  &  \bf{Features}  &  \bf{Algorithms Used}  &  \bf{Accuracy}  &  \bf{Precision}  &  \bf{Recall}  &  \bf{F1-score}  \\ \midrule
\citep{3}  &  $ 18 $  &  ExtraTrees, RF,  &  $99.88\%$  &  ---  &  ---  &  ---  \\
                      &          &  Logistic Regression,  &  &  &  &  \\
                      &          &  XGB  &  &  &  &  \\
                      &          &    &  &  &  &  \\
\citep{6}  &  $ 24 $  &  SVM, MPL, RF,  &  $95.16\%$  &  ---  &  ---  &  $93.80\%$  \\
                      &          &  Naïve Bayes,  &  &  &  &  \\
                      &          &  Bayesian network  &  &  &  &  \\
                      &          &    &  &  &  &  \\
\citep{7}  &  $ 24 $  &  Random Forest  &  $91.67\%$  &  ---  &  ---  &  ---  \\
\citep{4}  &  $ 14 $  &  CR  &  $84.21\%$  &  $83.70\%$  &  $84.20\%$  &  $83.70\%$  \\
\citep{5}  &  $ 15 $  &  SVM, RF, DT  &  $84.00\%$  &  ---  &  ---  &  $92.00\%$  \\
\citep{8}  &  $ 32 $  &  SVM  &  $81.48\%$  &  ---  &  ---  &  --- \\ \midrule
Proposed Model  &  $ 25 $  &  Stacked Ensemble  &  $100\%$  &  $100\%$  &  $100\%$  &  $100\%$  \\
                      &          &  Machine Learning  &  &  &  &  \\  \bottomrule
\end{tabular}}
\label{table_4.2}
\end{table}

Thus, the proposed highly accurate \& risk-free Stacked Ensemble Machine Learning model is capable of classifying individuals as COVID-19 +ve or -ve based on Hematological profile of their Blood-samples.

\section {Conclusion}
To combat the pandemic, it is imperative to ramp up the number of tests conducted in a country with high accuracy and precision for reliable screening of infected patients in a short time. To supplement the conventional time-consuming and scarce RT-PCR test this study proposes a risk-free `Stacked Ensemble Machine Learning' model to diagnose COVID-19 patients with relatively higher accuracy \& precision from their Hematological data obtained from routine blood tests.

This study characterized the limitations, risks associated with the models having unknown or less precision-recall in case of COVID-19 diagnosis exploiting the R-curve and maximum number of infections from simple SIR infectious disease model generalized for COVID-19 and hence harnessed the versatility-robustness of stacked ensemble model to resolve the problem. Nonetheless, this study keeps future scope of further development to address some challenges including usage of huge-diverse-high quality dataset, rigorous public testing and validation, incorporation of more attributes, and multiclass classification other than binary classification performed in this work.


\bigskip
\end{document}